%% file: main.tex
\documentclass[10pt,twocolumn,letterpaper]{article}

\usepackage[algorithms]{wacv}     

\definecolor{wacvblue}{rgb}{0.21,0.49,0.74}
\usepackage[pagebackref,breaklinks,colorlinks,allcolors=wacvblue]{hyperref}

\def\wacvPaperID{} 
\def\confName{WACV}
\def\confYear{2026}

\newcommand{\name}{SMBlurDetect \xspace}

\usepackage{multirow}
\usepackage{algorithm}
\usepackage{algpseudocode}
\usepackage{amsmath}
\usepackage{amssymb}
\usepackage{cases} 

\title{Subtle Motion Blur Detection and Segmentation from Static Image Artworks}

\author{
Ganesh Samarth$^{*}$\\
Amazon Prime Video\\
{\tt\small gxnesh@amazon.com}
\and
Sibendu Paul$^{*}$\\
Amazon Prime Video\\
{\tt\small sibendu@amazon.com}
\and
Solale Tabarestani$^{*}$\\
Amazon Prime Video\\
{\tt\small solaleta@amazon.com}
\and
Caren Chen\\
Amazon Prime Video\\
{\tt\small carechen@amazon.com}
}

\begin{document}
\maketitle
\begingroup
\renewcommand\thefootnote{\fnsymbol{footnote}}
\footnotetext[1]{* Primary authors with equal contribution.}
\endgroup

\input{section/abstract}
\input{section/intro}

\input{section/related}
\input{section/methods}
\input{section/exp}

\clearpage
\bibliographystyle{ieeenat_fullname}
\bibliography{blur}

\end{document}

%% file: section/abstract.tex
\begin{abstract}
Streaming services serve hundreds of millions of viewers worldwide, where visual assets—thumbnails, box art, and cover images—play a critical role in capturing attention and driving engagement. Motion blur, particularly subtle blur that is difficult to detect at first glance, remains one of the most pervasive quality degradation issues affecting these artworks, diminishing visual clarity and negatively impacting user trust and click-through rates. Despite its importance, motion blur detection from static images remains underexplored, with existing methods and datasets focusing on severe, visually obvious blur while lacking the fine-grained, pixel-level annotations needed for quality-critical applications. Popular benchmarks like GOPRO and NFS are dominated by strong synthetic blur unsuitable for detecting subtle, localized artifacts, and their "sharp" reference images often contain residual blur, creating ambiguous training signals. To address this gap, we propose SMBlurDetect, a unified framework combining high-quality motion-blur–specific dataset generation with an end-to-end robust detector capable of zero-shot detection at multiple granularities. Our dataset pipeline synthesizes realistic motion blur from super-high-resolution aesthetic images using controllable camera- and object-motion simulations over SAM-segmented regions, enhanced with alpha-aware compositing and balanced sampling to create subtle, spatially targeted blur with precise ground-truth masks. Leveraging this data, we train a U-Net–based detector with ImageNet-pretrained encoders through a hybrid mask- and image-centric strategy, incorporating progressive curriculum learning, hard negatives, focal loss, blur-frequency channels, and resolution-aware augmentation. Our approach achieves strong zero-shot generalization, significantly outperforming dataset-specific supervised baselines with 89.68\% accuracy on GoPro (vs 66.50\% baseline) and 59.77\% Mean IoU on CUHK (vs 9.00\% baseline), demonstrating 6.6$\times$ improvement in segmentation. Qualitative results on internal subtle-blur samples show accurate localization of imperceptible blur artifacts on critical semantic regions such as faces and hands, enabling automated filtering of low-quality frames during artwork generation and supporting precise region-of-interest extraction for intelligent cropping—ultimately ensuring consistently sharp, appealing imagery that enhances customer satisfaction and maintains the premium visual standards essential for streaming services success.

\end{abstract}

%% file: section/intro.tex
\section{Introduction}
\label{sec:intro}

Streaming services today serve hundreds of millions of viewers worldwide, offering vast catalogs of movies, series, and short-form videos. In this highly competitive discovery landscape, visual assets—thumbnails and cover images—play a critical role in capturing attention, driving engagement, and influencing viewer choice. Because these artworks are often extracted directly from in-video frames, they are susceptible to various forms of quality degradation, with motion blur being one of the most common and impactful. Even subtle blur, often difficult to detect at first glance, can diminish visual clarity, reduce perceived production quality, and negatively affect user trust and click-through rates. Moreover, motion blur is not limited to automated extractions; artworks submitted by operators or third-party vendors can also contain subtle blur artifacts that are easy to overlook yet detrimental to overall presentation quality. Detecting such blur in static images is inherently challenging—motion is a temporal phenomenon, and distinguishing subtle blur from natural softness or compression artifacts requires sophisticated analysis. Despite these challenges, robust motion blur detection is essential. It enables the filtering of low-quality frames during artwork generation, supports accurate region-of-interest (ROI) extraction, improves intelligent cropping and resizing, and strengthens downstream visual understanding tasks. As streaming services continue to scale personalized and high-impact visual experiences, reliable and precise blur detection becomes a foundational quality-control component—ensuring consistently sharp, appealing, and engaging imagery for every viewer. 

\begin{figure}[h!]
    \centering
    
    \begin{subfigure}[t]{0.45\linewidth}
        \centering
        \includegraphics[width=\linewidth]{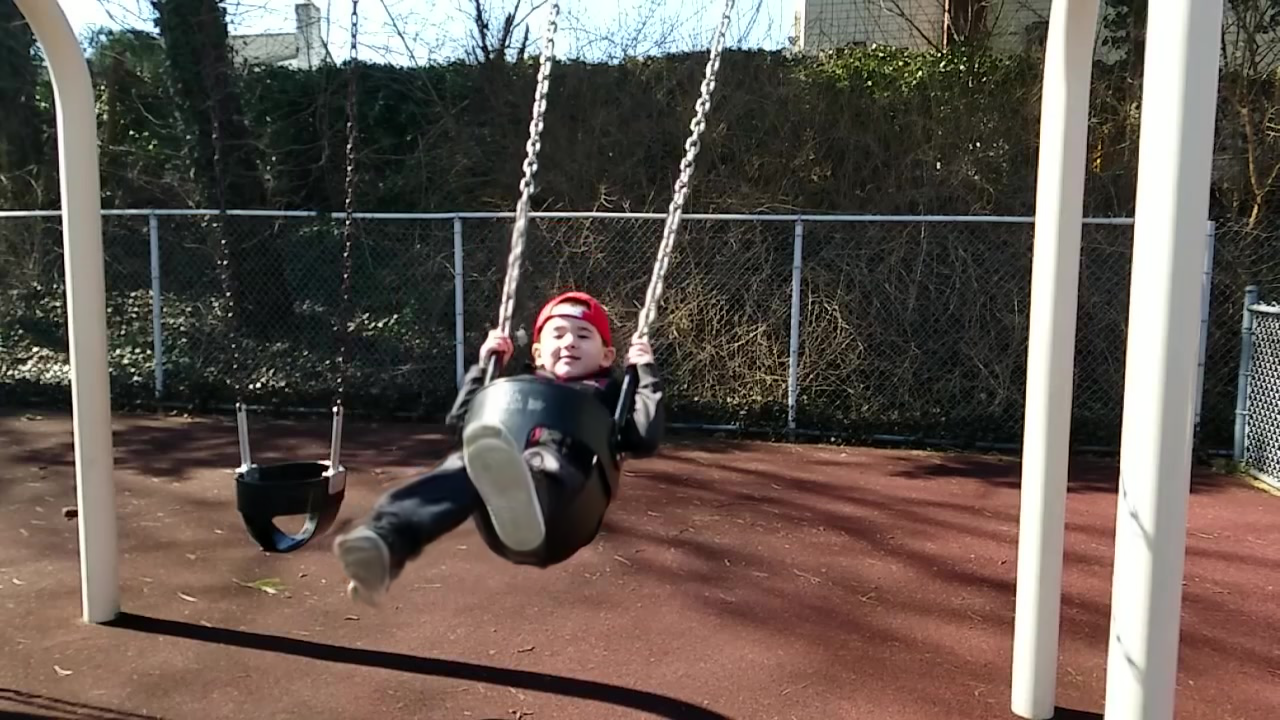}
        \caption{}
    \end{subfigure}
    \hfill
    \begin{subfigure}[t]{0.45\linewidth}
        \centering
        \includegraphics[width=\linewidth]{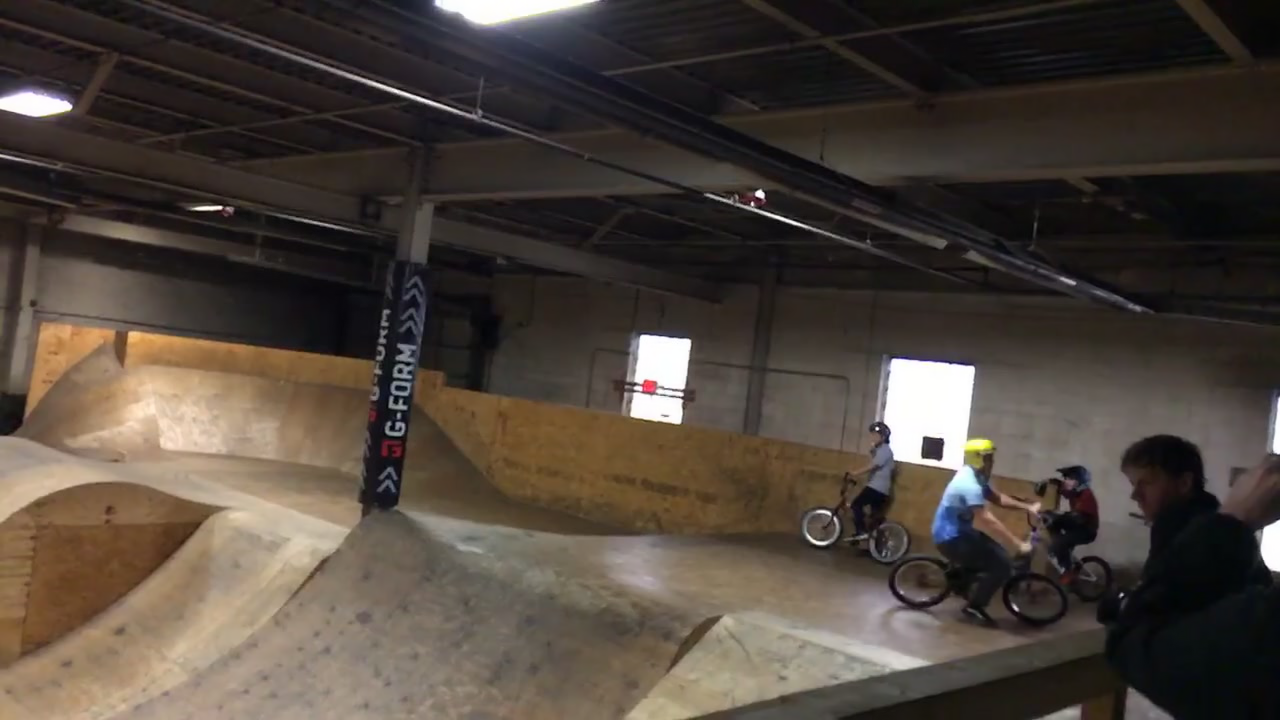}
        \caption{}
    \end{subfigure}


    
    \caption{Examples from GOPRO~\cite{gopro} and NFS~\cite{nfs} where “sharp” images still contain subtle motion blur, illustrating why existing datasets are unreliable for training motion-blur detectors for quality-critical applications.}
    \label{fig:blur_examples}
\end{figure}

Despite its importance, motion blur detection from a single static image remains underexplored, especially at the fine-grained level required for quality-critical applications such as streaming service artworks. Existing methods and datasets overwhelmingly target severe, visually obvious blur, with popular benchmarks like GOPRO~\cite{gopro}, NFS~\cite{nfs}, and DVD~\cite{Kupyn_2019_ICCV} dominated by strong synthetic motion blur generated through frame averaging or simulated camera shake. While suitable for general deblurring research, these conditions fail to represent the subtle, localized blur frequently observed in real production frames. Many blur datasets~\cite{chuk} further exhibit a strong bias toward defocus blur — making them ill-suited for training detectors specifically sensitive to motion blur. Furthermore, the so-called “sharp” reference images in these datasets often contain residual, low-level blur (also shown in Figure~\ref{fig:blur_examples}), creating ambiguous training signals that hinder models from learning to reliably separate truly sharp content from subtly motion-blurred images. 
Moreover, most existing approaches provide only global image-level predictions, lacking the patch-level granularity needed to identify blur affecting small but critical semantic regions such as faces or hands. The absence of high-quality, high-volume, pixel-level motion-blur masks further restricts progress, leaving current state-of-the-art methods inadequate for detecting and localizing the subtle motion blur artifacts that significantly degrade the visual quality of streaming artworks.

To address this gap, we propose \name, a unified framework that integrates high-quality, motion-blur–specific dataset generation with an end-to-end, robust motion-blur detector. Our approach supports zero-shot detection at multiple granularities and produces accurate blur-mask segmentation suitable for quality-critical applications. Our dataset pipeline synthesizes realistic motion blur from super-high-resolution aesthetic images using controllable camera- and object-motion simulations over SAM-segmented regions, enhanced with alpha-aware compositing, photometric effects, and balanced sampling to create subtle, spatially targeted blur along with precise ground-truth masks. Leveraging this dataset, we train a U-Net–based detector with ImageNet-pretrained encoders through a hybrid mask- and image-centric strategy, incorporating progressive difficulty, hard negatives, focal loss, a blur-frequency input channel, and resolution-aware augmentation to accurately localize subtle blur. This scalable generation-and-detection pipeline delivers strong zero-shot generalization, outperforming dataset-specific supervised baselines, with quantitative classification results on GoPro and NFS and qualitative segmentation performance on both internal subtle-blur samples and an additional benchmark dataset.

Overall, our contributions can be summarized as follows:

\begin{itemize}
\item To the best of our knowledge, we are the first to highlight critical limitations in existing motion-blur research and datasets. Prior work predominantly targets severe motion blur or defocus blur and lacks high-quality, pixel-level annotations, making it insufficient for detecting the subtle, localized motion blur that impacts artwork quality on streaming services.
\item We introduce a high-quality motion-blur dataset generation pipeline that applies controllable camera- and object-motion simulations to SAM-segmented regions, producing realistic, subtle, and spatially targeted motion blur along with precise ground-truth masks.
\item We present an end-to-end motion-blur localization model built on a U-Net backbone that integrates hybrid mask- and image-centric training, progressive difficulty scheduling, hard-negative mining, focal loss, a blur-frequency channel, and resolution-aware augmentation. Together, these components enable robust and accurate detection of subtle, localized motion blur.
\item We conduct extensive quantitative and qualitative evaluations, showing strong classification performance on GoPro and NFS, effective segmentation on an additional benchmark, and superior zero-shot generalization on internal subtle-blur datasets.
\end{itemize}

%% file: section/related.tex
\section{Related Work}
\label{sec:related}
\subsection{Blur Detection Methods}

Early blur detection relied on hand-crafted features. Shi et al.~\cite{chuk} proposed discriminative features across gradient, frequency, and learned domains, introducing the CUHK dataset with pixel-wise annotations. Chakrabarti et al.~\cite{chakrabarti2010analyzing} developed local blur cues using Gaussian Scale Mixture models, while Golestaneh and Karam~\cite{alireza2017spatially} proposed HiFST using multiscale DCT coefficients for type-agnostic detection.

Deep learning transformed blur detection. Sun et al.~\cite{sun2015learning} pioneered CNN-based spatially-varying blur kernel estimation. Ma et al.~\cite{ma2018deep} demonstrated that high-level semantic features outperform low-level cues in Deep Blur Mapping, achieving 0.853 ODS F-score. Tang et al.~\cite{tang2019defusionnet} introduced DeFusionNET with recurrent multi-scale fusion, setting state-of-the-art results. For motion-specific detection, Gong et al.~\cite{gong2017motion} developed end-to-end FCN mapping to dense motion flow fields. Kim et al.~\cite{kim2018defocus} proposed multi-scale CNNs for joint defocus and motion blur detection, though without type discrimination. Zhang et al.~\cite{zhang2018learning} unified blur estimation with quality classification in ABC-FuseNet, achieving 0.869 average precision on CUHK.

\subsection{Motion Blur Datasets and Domain Gap}

The GoPro dataset~\cite{gopro} by Nah et al. synthesized blur by averaging consecutive 240fps frames, becoming the standard benchmark with 3,214 pairs. Su et al.~\cite{Kupyn_2019_ICCV} introduced DVD with 6,708 video pairs. The CUHK dataset~\cite{chuk} provides 1,000 images with pixel-wise annotations for both blur types. The REDS dataset~\cite{nah2019ntire} offers 30,000 frames with compression artifacts for practical evaluation.

However, these datasets have critical limitations. Rim et al.~\cite{rim2020real} demonstrated that GoPro-trained models fail on real blur due to synthetic-to-real domain gap, introducing RealBlur with beam-splitter-captured ground truth. More fundamentally, the ``sharp'' reference images in GoPro and similar datasets often contain residual blur creating ambiguous training signals. Datasets with dense pixel-level motion blur annotations remain scarce, and existing work predominantly targets severe, visually obvious blur rather than the subtle artifacts critical for quality-sensitive applications.

\subsection{Image Deblurring and Blur Synthesis}

Deblurring methods reveal blur-relevant features. Nah et al.'s DeepDeblur~\cite{gopro} established end-to-end multi-scale CNNs achieving 29.08 dB on GoPro. Kupyn et al.~\cite{kupyn2018deblurgan} introduced adversarial training with perceptual loss in DeblurGAN. Zhang et al.~\cite{zhang2019deep} proposed hierarchical multi-patch processing in DMPHN, achieving real-time performance relevant for localization tasks. Recent transformer approaches achieved state-of-the-art results: Zamir et al.'s MPRNet~\cite{mehri2021mprnet} introduced supervised attention modules learning per-pixel reweighting, Restormer~\cite{zamir2022restormer} used transposed attention for 32.92 dB, and Chen et al.'s NAFNet~\cite{na2024efficiency} demonstrated activation-free networks achieving 33.69 dB with 8.4\% computational cost.

Realistic blur synthesis is critical for robust training. Brooks and Barron~\cite{brooks2019learning} introduced differentiable blur synthesis handling complex motions.  Boracchi and Foi~\cite{boracchi2012modeling} established statistical foundations for motion kernel generation from camera shake.

%% file: section/methods.tex
\section{Method Overview}
\label{sec:method}

\begin{figure*}[t]
\centering
\includegraphics[width=\linewidth]{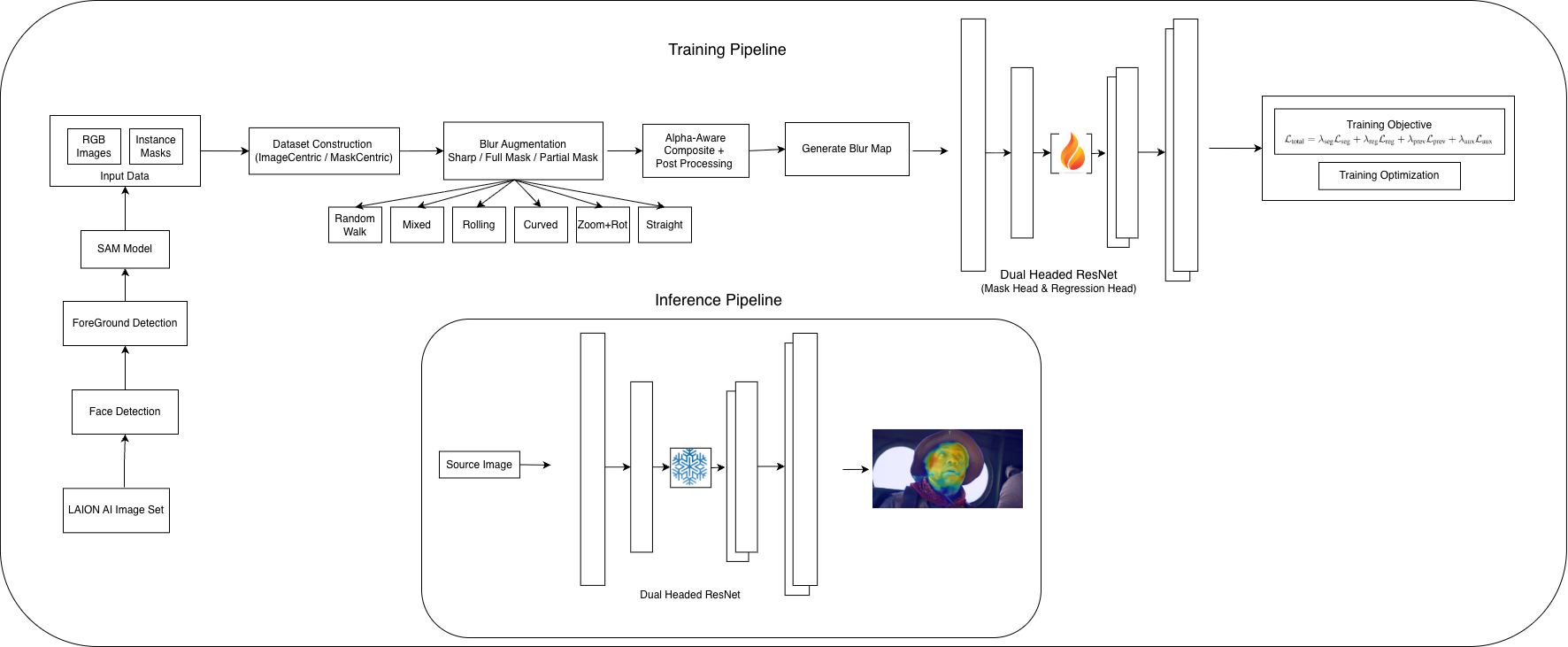}
\caption{End-to-end SMBlurDetect Pipeline Overview. The system comprises three main components: (1) \textbf{Dataset Preparation:} High-quality images from LAION-5B~\cite{laion} are processed through SAM-based segmentation to extract foreground instance masks for critical regions (faces, hands, hair), with hybrid Mask-Centric and Image-Centric sampling strategies controlled by $mask_{ratio}$ parameter. (2) \textbf{Motion Blur Synthesis:} Six physically motivated blur types (straight, curved, zoom with rotation, random-walk, edge-ring, rolling) are applied using exposure-based temporal integration and PSF convolution, with alpha-aware compositing and adaptive edge feathering to generate photorealistic blur with precise ground-truth masks. (3) \textbf{Dual-Head U-Net Architecture:} A ResNet-50 ImageNet-pretrained encoder with decoder skip connections produces binary blur segmentation (Mask Head) and continuous blur-intensity maps (Regression Head), trained using composite loss through three-stage progressive curriculum learning, enabling accurate multi-granularity detection of subtle motion blur in artworks.}
\label{fig:pipeline}
\end{figure*}

We present an end-to-end training pipeline (as shown in Figure~\ref{fig:pipeline}) for motion-blur segmentation that synthesizes photorealistic blur on instance masks and trains a dual-head segmentation network to predict both binary blur-region masks and continuous blur-intensity maps. Motivated by the scarcity of high-quality, mask-level motion blur annotations—and the difficulty of capturing subtle blur that is often missed in existing state-of-the-art datasets—our approach generates physically grounded synthetic blur examples to enrich training data. Combined with a carefully designed training objective, this pipeline enables robust learning of both the presence and severity of motion blur, overcoming the limitations of current real-world datasets.

\begin{figure}[t]
\centering
\includegraphics[width=\linewidth]{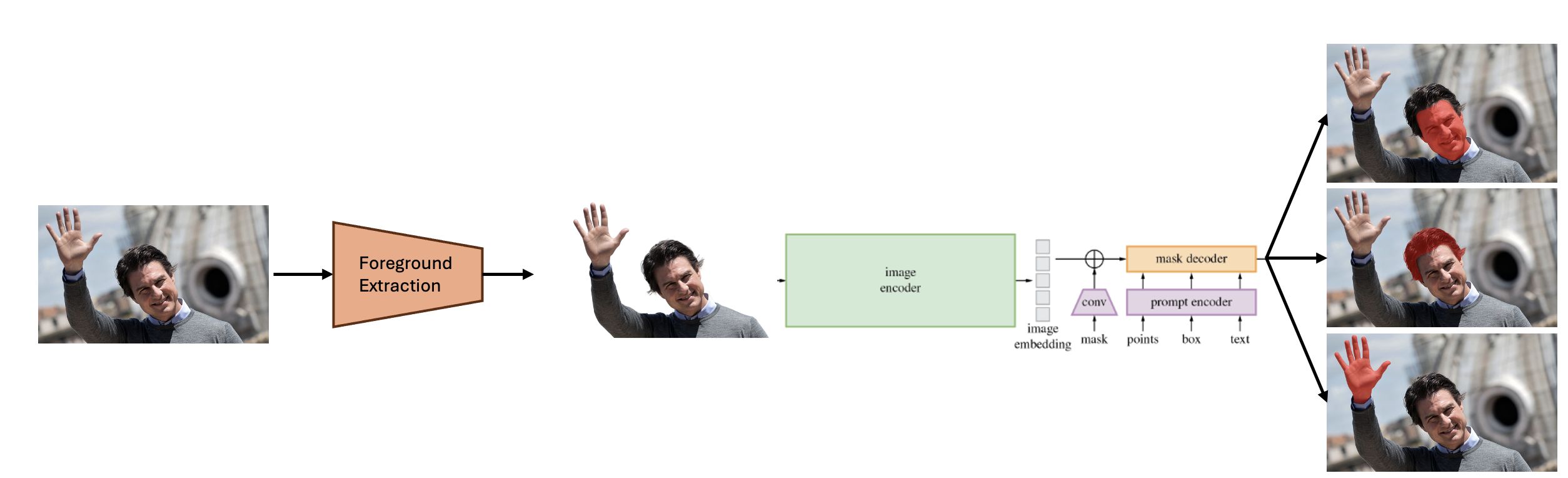}
\caption{Foreground instance masks highlighting regions most susceptible to motion blur, for example face, hair, and hands.}
\label{fig:dataset_gen}
\end{figure}

\subsection{Dataset Preparation}
We begin with high-quality, aesthetically pleasing images filtered from LAION-5B~\cite{laion}, further restricting the dataset to images containing human faces with high confidence to closely match our use case. For each selected RGB image, we extract precise foreground instance masks using a combination of foreground extraction and the Segment Anything Model (SAM)~\cite{sam2023}. Because even subtle motion blur on foreground actors can significantly degrade user experience, we focus exclusively on modeling blur within the foreground region. The resulting instance masks are stored as grayscale images, with foreground pixels assigned a value of 255 and background pixels assigned 0. To ensure reproducibility and prevent data leakage, we generate train and validation splits through seeded shuffling. As illustrated in Figure~\ref{fig:dataset_gen}, we segment fine-grained instances from the actor (present in foreground) such as hands, hair, and face that are especially susceptible to motion blur.

\begin{figure}[t]
\centering
\includegraphics[width=\linewidth]{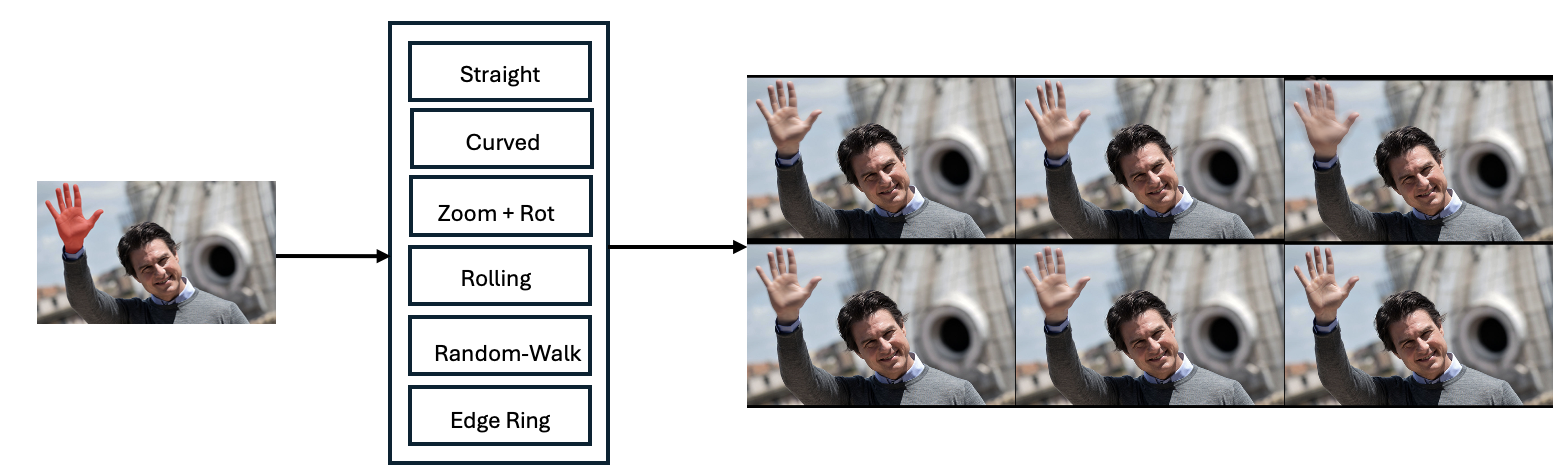}
\caption{Our photorealistic blur augmentation pipeline implements six distinct blur types, each designed to model a specific motion scenario. For every instance, one blur type is applied at varying strengths, enabling multiple motion patterns to coexist within a single image. 
}
\label{fig:blur_augmentation}
\end{figure}

\textbf{Dataset Construction Strategies.} We employ two complementary dataset construction approaches to balance robustness and localization. 
\begin{itemize}
    \item \textbf{Mask-Centric approach:} Constructs training samples as (image, single-mask) pairs. This approach emphasizes small or localized blur regions and effectively increases dataset diversity by treating each instance independently, as shown in Figure~\ref{fig:blur_augmentation}. Additionally, randomized crop scales and aspect ratios further improve generalization.
    
    \item \textbf{Image-Centric approach:} Applies blur at the full-image level, sampling entire images with varied backgrounds and scenes—including cases with varying or no blur. Multiple instance masks within each image can be selected for augmentation, with each mask receiving an independently chosen blur type. This strategy improves robustness to diverse contexts and prevents overfitting to always-present blur patterns or overly narrow instance-level cues.
    
\end{itemize}

To combine the strengths of both approaches, we adopt a hybrid sampling strategy during training, controlled by a $mask\_{ratio}$ parameter. This allows us to balance scene-level diversity with instance-level precision, enabling the model to learn both global context and fine-grained blur localization.

\subsection{Motion Blur Synthesis}
Rather than applying uniform blur or predefined blur kernels across instances, we aim to synthesize photorealistic motion blur that mimics real-world camera and object motion patterns. The motion blur augmentation module of \name implements physically motivated blur types through two primary mechanisms: exposure-based temporal integration and point spread function (PSF) convolution.

\textbf{Blur Type Categories.}
Our augmentation pipeline implements six motion-specific blur types designed to capture a wide spectrum of real-world motion patterns, as illustrated in Figure~\ref{fig:blur_augmentation}.
\begin{itemize}
    \item \textbf{Straight:} Simulates linear camera or object motion through exposure-based temporal integration. Multiple warped frames of the premultiplied foreground are accumulated along a straight trajectory to generate classic motion streaks.
    \item \textbf{Curved:} Models smooth motion along Bezier-curve trajectories. Warped frames are temporally averaged to produce natural, curved blur trails reflecting non-linear motion paths.
    \item \textbf{Zoom with Rotation:} Combines mild rotational drift with scale variation to mimic camera instability or object rotation during exposure.
    \item \textbf{Random-Walk:} Uses PSF-based convolution with a stochastic motion kernel to capture jittery, shake-like movement. The irregular PSF path is convolved with the premultiplied foreground to produce realistic shake blur.
    \item \textbf{Edge-Ring:} Applies motion blur selectively within a thin band around object boundaries, capturing the characteristic edge smear common in fast-moving objects where contours blur more strongly than interiors.
    \item \textbf{Rolling:} Extends straight-motion blur by applying row-dependent positional offsets, replicating rolling-shutter distortions typical of CMOS sensors.
\end{itemize}

\textbf{Blur Type Selection and Application.} For each selected instance mask (whether from Image-Centric or Mask-Centric sampling), we independently choose which blur augmentation to apply from the six types described above. To maximize training diversity and realism, we employ several randomization strategies:

\begin{itemize}
    \item \textbf{Blur Coverage Modes:} Each instance is randomly assigned one of three coverage modes: (1) \textit{sharp} (no blur applied), (2) \textit{full-mask} (blur applied uniformly across the entire instance), or (3) \textit{partial-mask} (blur applied only to randomly selected subregions). This enables the model to learn both fully-blurred and partially-blurred instances, as well as sharp regions for negative examples.
    
    \item \textbf{Mixed Blur Types:} In Image-Centric construction, where multiple instance masks exist within a single training image, each instance independently receives a randomly selected blur type. This \textit{mixed mode} allows diverse motion patterns to coexist within one image—for example, a face exhibiting straight blur while hands show curved blur—better reflecting real-world scenarios where different objects undergo different motions simultaneously.
\end{itemize}

We apply alpha-aware compositing with adaptive edge feathering when blending the blurred foreground onto the background. This prevents hard seams at motion-blur boundaries, which would otherwise create unrealistic artifacts. Without this smoothing, the model could mistakenly learn to detect sharp mask edges (i.e., perform edge detection) instead of true motion-blur segmentation, since a hard transition between blurred and non-blurred regions is not physically realistic.

\subsection{\name\ Model Architecture}
We employ a U-Net architecture~\cite{ronneberger2015u} with a ResNet-50 encoder pre-trained on ImageNet, implemented using the PyTorch Segmentation Models (SMP) library. The decoder features skip connections from encoder layers to recover spatial details lost during downsampling.

To jointly model blur localization and severity, we employ a dual-head design: 
\begin{itemize}
    \item \textbf{Mask Head:} A 1×1 convolution layer producing $mask_{logits}$ (no activation). After applying a sigmoid, these logits form a probabilistic motion-blur segmentation map.
    \item \textbf{Regression Head:} A 1×1 convolution followed by sigmoid activation that directly predicts a continuous blur-intensity map ($reg_{map} \in [0,1]$).
\end{itemize}

We further apply deep supervision by attaching auxiliary prediction heads to intermediate decoder stages, adding multi-scale loss terms that encourage stable feature learning and improve convergence.

\subsection{Training Objective}
We employ a composite loss function $\mathcal{L}_{\text{total}}$ for motion blur localization that balances multiple objectives to ensure accurate boundary localization, continuous intensity prediction, and training stability.

\begin{equation}
\mathcal{L}_{\text{total}} = \lambda_{\text{seg}} \mathcal{L}_{\text{seg}} + \lambda_{\text{reg}} \mathcal{L}_{\text{reg}} + \lambda_{\text{prev}} \mathcal{L}_{\text{prev}} + \lambda_{\text{aux}} \mathcal{L}_{\text{aux}}
\end{equation}

\noindent where $\lambda_{\text{seg}}$, $\lambda_{\text{reg}}$, $\lambda_{\text{prev}}$, and $\lambda_{\text{aux}}$ are weighting coefficients balancing the contribution of each component.

\subsubsection{Segmentation Loss Component}
The segmentation loss combines three complementary terms: Binary Cross-Entropy (BCE), Dice loss~\cite{li2020dice}, and Focal Tversky loss~\cite{abraham2019novel} on $mask_{logits}$ versus $y_{mask}$. We use BCE loss to enforce pixel-level accuracy and well-calibrated blur probabilities. The Dice loss handles class imbalance and encourages precise boundary localization, while Focal Tversky loss emphasizes hard examples and small blur regions.

{
\small
\begin{equation}
\mathcal{L}_{\text{seg}} = \mathcal{L}_{\text{BCE}}(\hat{y}_{\text{mask}}, y_{\text{mask}}) + \mathcal{L}_{\text{Dice}}(\hat{y}_{\text{mask}}, y_{\text{mask}}) + \mathcal{L}_{\text{FT}}(\hat{y}_{\text{mask}}, y_{\text{mask}})
\end{equation}
}

\noindent where $\hat{y}_{\text{mask}}$ (or $\hat{y}_i$) is the predicted blur probability map after sigmoid activation, and $y_{\text{mask}}$ ($y_i$) is the binary ground truth mask. The individual terms are defined as:\footnote{$\hat{y}_{\text{mask}}$ and $y_{\text{mask}}$ are represented as $\hat{y}_i$ and $y_i$ respectively in the following equations.} \\

Binary Cross-Entropy Loss:
\begin{equation}
\mathcal{L}_{\text{BCE}} (\hat{y}_i, y_i) = -\frac{1}{N}\sum_{i=1}^{N} \left[ y_i \log(\hat{y}_i) + (1-y_i)\log(1-\hat{y}_i) \right] \\
\end{equation}

Dice Loss:
\begin{equation}
\mathcal{L}_{\text{Dice}}(\hat{y}_i, y_i)  = 1 - \frac{2\sum_{i=1}^{N} \hat{y}_i y_i + \epsilon}{\sum_{i=1}^{N} \hat{y}_i + \sum_{i=1}^{N} y_i + \epsilon} \\
\end{equation}

Focal Tversky Loss:
\begin{equation}
\mathcal{L}_{\text{FT}} = \left(1 - \text{TI}\right)^{\gamma}
\end{equation}

\noindent where the Tversky Index is:
{\footnotesize
\begin{equation}
\text{TI}(\hat{y}_i, y_i)  = \frac{\sum_{i=1}^{N} \hat{y}_i y_i + \epsilon}{\sum_{i=1}^{N} \hat{y}_i y_i + \alpha\sum_{i=1}^{N} (1-\hat{y}_i) y_i + \beta\sum_{i=1}^{N} \hat{y}_i (1-y_i) + \epsilon}
\end{equation}
}

\noindent with hyperparameters $\alpha$ and $\beta$ controlling false negative and false positive penalties, $\gamma$ focusing on hard examples, and $N$ is the total number of pixels.

\subsubsection{Regression Loss Component}

We apply Huber loss between $reg_{map}$ and $y_{reg}$, masked by $y_{mask}$ to focus regression only within blurred regions. Huber loss provides robustness to outliers while preserving smooth intensity gradients.

\begin{equation}
\mathcal{L}_{\text{reg}} = \frac{1}{|\mathcal{M}|}\sum_{i \in \mathcal{M}} \mathcal{H}(\text{reg}_{\text{map},i}, y_{\text{reg},i})
\end{equation}

\noindent where $\mathcal{M} = \{i : y_{\text{mask},i} > 0\}$ denotes pixels within the ground truth blur mask, $\text{reg}_{\text{map},i}$ is the predicted continuous blur intensity, $y_{\text{reg},i}$ is the ground truth continuous blur intensity, and $\mathcal{H}$ is the Huber loss:
{
\small
\begin{equation}
\mathcal{H}(\text{reg}_{\text{map}}, y_{\text{reg}}) = \begin{cases}
\frac{1}{2}(\text{reg}_{\text{map}} - y_{\text{reg}})^2 & \text{if } |\text{reg}_{\text{map}} - y_{\text{reg}}| \leq \delta \\
\delta \left(|\text{reg}_{\text{map}} - y_{\text{reg}}| - \frac{1}{2}\delta\right) & \text{otherwise}
\end{cases}
\end{equation}
}

\noindent with $\delta$ controlling the transition from quadratic to linear behavior.

\subsubsection{Regularization Component}

To prevent degenerate all-zero or all-one predictions, we penalize extreme prediction distributions:

\begin{equation}
\mathcal{L}_{\text{prev}} = \left(\frac{1}{N}\sum_{i=1}^{N} \hat{y}_{\text{mask},i} - p_{\text{target}}\right)^2
\end{equation}

\noindent where $p_{\text{target}}$ is the expected prevalence of blur in the training distribution.

\subsubsection{Auxiliary Deep Supervision Loss Component}

To stabilize training and enhance motion-blur–specific feature learning, we apply auxiliary losses at intermediate decoder stages, providing additional gradient signals that guide the network more effectively.

\begin{equation}
\mathcal{L}_{\text{aux}} = \frac{1}{L}\sum_{l=1}^{L} \left[\mathcal{L}_{\text{BCE}}(\hat{y}_{\text{mask}}^{(l)}, y_{\text{mask}}^{(l)}) + \mathcal{L}_{\text{Dice}}(\hat{y}_{\text{mask}}^{(l)}, y_{\text{mask}}^{(l)})\right]
\end{equation}

\noindent where $\hat{y}_{\text{mask}}^{(l)}$ and $y_{\text{mask}}^{(l)}$ denote predictions and downsampled targets at scale $l$.

This multi-component objective stabilizes boundary predictions through Dice and Focal Tversky losses while preserving fine-grained blur intensity structure through the masked Huber regression.

\subsection{Training Optimization}

We optimize the model using AdamW with gradient clipping to prevent exploding gradients, employing optional mixed precision training (AMP) for computational efficiency and Distributed Data Parallel (DDP) for multi-GPU scaling. To prevent early training instability from overly complex augmentations, we adopt a three-stage curriculum learning schedule:

\begin{itemize}
    \item \textbf{Stabilization Stage:} Emphasizes straight and random-walk blur types to establish basic motion blur recognition.
    \item \textbf{Mixed Realism Stage:} Introduces curved, rolling, and edge-ring blur patterns to expand the model's understanding of diverse motion patterns.
    \item \textbf{Advanced Stage:} Enables zoom with rotation and mixed-instance scenarios for maximum realism and diversity.
\end{itemize}

This progressive training strategy allows the model to gradually learn from simple to complex blur patterns, improving convergence stability and final performance.

%% file: section/exp.tex
\section{Experiments}
\label{sec:exp}
In this section, we present both quantitative and qualitative results. For quantitative evaluation, we first report performance on two motion-blur–specific binary classification datasets: GoPro~\cite{gopro} and NFS~\cite{nfs}. We then assess motion-blur segmentation using the only available dataset with blur masks, CUHK~\cite{chuk}, despite its limited size. Finally, we provide qualitative visualizations to further analyze model behavior and highlight strengths and limitations.

\subsection{Evaluation Datasets}
To evaluate how well the trained model distinguishes motion-blurred images from sharp ones, we use two high-quality, state-of-the-art datasets commonly employed in image deblurring research: GoPro~\cite{gopro} and NFS~\cite{nfs}. The GoPro dataset~\cite{gopro} is constructed using a GoPro Hero4 Black camera by capturing 240 fps video and averaging 3–21 successive latent frames to synthesize motion blur of varying strengths. Each blurred or sharp image has a resolution of $1280\times720$. The NFS dataset adopts a similar methodology to generate real-world blurred and sharp image pairs. GoPro contains 8,800 training and 4,400 testing images, while NFS provides approximately 17,000 training and 3,000 testing images.

It is important to note that our \name\ model is trained exclusively on our custom dataset using the proposed photorealistic motion-blur augmentation pipeline, shown in Figure~\ref{fig:blur_augmentation}. We apply no additional fine-tuning on GoPro or NFS, and instead evaluate model performance in a zero-shot setting on these datasets.

For motion-blur segmentation, we use the CUHK dataset~\cite{chuk}, which includes both out-of-focus and motion-blurred images with corresponding human-labeled ground-truth blur regions. To align with our task, we filter the dataset to retain only motion-blurred examples, resulting in approximately 250 images (out of the original 1,000) with mask-level annotations. 

\subsection{Evaluation Metrics}
We evaluate model performance using both pixel-level segmentation and image-level binary classification metrics, depending on the dataset characteristics and available ground-truth annotations.

For \textbf{pixel-level segmentation} on the CUHK dataset, which provides dense pixel-wise blur annotations, we report Pixel Accuracy, Mean IoU (Intersection over Union), Weighted IoU, and class-wise Precision, Recall, and F1 Score. These metrics are computed at a threshold of $\tau = 0.5$ on the predicted probability maps. To characterize the full precision–recall trade-off, we additionally compute precision–recall curves across multiple thresholds and report the area under the curve (AUC).

For \textbf{image-level binary classification} on GoPro and NFS datasets, where labels indicate whether an image contains motion blur, we evaluate using Overall Accuracy, Precision, Recall (Sensitivity), Specificity, F1 Score, and ROC AUC. These metrics assess the model's ability to discriminate between blurred and sharp images based on aggregated predictions.

To ensure fair comparison across models and configurations, the optimal decision threshold $\tau^*$ is determined separately for each method via grid search over the validation set, selecting the threshold that maximizes IoU for segmentation tasks or F1 score for classification tasks.

\subsection{Baseline Method}

We compare our approach against the blur segmentation and classification method by Kim et al.~\cite{kim2018defocus}. To the best of our knowledge, there is no established baseline that specifically targets \emph{motion blur segmentation}; most related works instead focus on deblurring or generic blur detection/estimation without producing motion-blur-specific segmentation masks. Kim et al. is the closest directly comparable prior work, as it explicitly segments and distinguishes motion blur (alongside defocus blur) using a multi-scale CNN with contextual feature extraction and patch-based classification, combining hand-crafted and learned deep features. While effective, their fixed-scale patch processing and limited blur-type modeling can struggle when blur extent and intensity vary significantly across the image.

\subsection{Quantitative Results}

\subsubsection{CUHK Dataset Performance}

Table~\ref{tab:cuhk_results} presents pixel-level segmentation results on the CUHK dataset. Our method significantly outperforms the baseline across all metrics, achieving a Mean IoU of 59.77\% compared to 9.00\% for the baseline method. The dramatic improvement in recall (73.81\% vs 9.04\%) demonstrates our model's superior ability to detect blurred regions, while maintaining competitive precision (75.86\% vs 96.37\%). The baseline's extremely high precision but low recall indicates a conservative prediction strategy that misses most blur regions. Our balanced performance (F1: 74.82\% vs 16.52\%) validates the efficacy of our synthetic blur augmentation and dual-head model architecture.

\begin{table}[h]
\centering
\caption{Pixel-level segmentation performance on CUHK dataset~\cite{chuk}}
\label{tab:cuhk_results}
\begin{tabular}{l|cc}
\hline
\textbf{Metric} & \textbf{Ours} & \textbf{Kim et al.~\cite{kim2018defocus}} \\
\hline
Pixel Accuracy (\%) & \textbf{72.69} & 47.11 \\
Mean IoU (\%) & \textbf{59.77} & 9.00 \\
Weighted IoU (\%) & \textbf{57.19} & 23.81 \\
Mean Class Accuracy (\%) & \textbf{72.56} & 54.28 \\
\hline
\multicolumn{3}{l}{\textit{Blur Class Performance:}} \\
\hline
Precision (\%) & 75.86 & \textbf{96.37} \\
Recall (\%) & \textbf{73.81} & 9.04 \\
F1 Score (\%) & \textbf{74.82} & 16.52 \\
IoU (Blur) (\%) & \textbf{59.77} & 9.00 \\
\hline
\end{tabular}
\end{table}

\subsubsection{GoPro Dataset Performance}

On the GoPro dataset~\cite{gopro}, our method achieves 89.68\% accuracy (zero-shot) with a well-balanced confusion matrix, as illustrated in Table~\ref{tab:gopro_results}. The baseline method exhibits severe overfitting toward the sharp class, achieving perfect specificity (100\%) but poor recall (33\%). This indicates the model's failure to generalize to motion-blurred images. In contrast, our approach maintains strong performance on both classes with an ROC AUC of 0.94, demonstrating robust discrimination capability. The 23.18\% improvement in overall accuracy and 53.32\% gain in blur detection recall highlight the advantages of our photorealistic motion-blur augmentation strategy.

\begin{table}[h]
\centering
\caption{Binary classification performance on GoPro dataset (2.2K sharp and 2.2K blur images).}
\label{tab:gopro_results}
\begin{tabular}{l|cc}
\hline
\textbf{Metric} & \textbf{Ours} & \textbf{Kim et al.~\cite{kim2018defocus}} \\
\hline
Accuracy (\%) & \textbf{89.68} & 66.50 \\
Precision (\%) & 92.54 & \textbf{100.00} \\
Recall (\%) & \textbf{86.32} & 33.00 \\
Specificity (\%) & 93.05 & \textbf{100.00} \\
F1 Score (\%) & \textbf{89.32} & 49.62 \\
ROC AUC & \textbf{0.94} & 0.79 \\
\hline
Sharp Accuracy (\%) & 93.05 & \textbf{100.00} \\
Blur Accuracy (\%) & \textbf{86.32} & 33.00 \\
\hline
\end{tabular}
\end{table}

\subsubsection{NFS Dataset Performance}

The performance of \name\ further confirms the superiority of our approach, as shown in Table~\ref{tab:nfs_results}. Our method achieves 80.33\% accuracy compared to 59.33\% for the baseline. The baseline again shows extreme bias toward sharp predictions (Specificity: 99.87\%, Recall: 18.80\%), correctly classifying almost no blurred images. Our balanced performance (Blur Accuracy: 73.93\%, Sharp Accuracy: 86.73\%) with an F1 score of 79\% demonstrates effective generalization to diverse motion patterns.

\begin{table}[h]
\centering
\caption{Binary classification performance on NFS dataset (1.5K sharp and 1.5K blur images).}
\label{tab:nfs_results}
\begin{tabular}{l|cc}
\hline
\textbf{Metric} & \textbf{Ours} & \textbf{Kim et al.~\cite{kim2018defocus}} \\
\hline
Accuracy (\%) & \textbf{80.33} & 59.33 \\
Precision (\%) & 84.79 & \textbf{99.30} \\
Recall (\%) & \textbf{73.93} & 18.80 \\
Specificity (\%) & 86.73 & \textbf{99.87} \\
F1 Score (\%) & \textbf{79.00} & 31.61 \\
ROC AUC & \textbf{0.85} & 0.70 \\
\hline
Sharp Accuracy (\%) & 86.73 & \textbf{99.87} \\
Blur Accuracy (\%) & \textbf{73.93} & 18.80 \\
\hline
\end{tabular}
\end{table}

\subsection{Analysis and Discussion}

Our experimental results demonstrate that \name\ achieves strong zero-shot generalization across diverse motion blur datasets, significantly outperforming supervised baselines despite being trained exclusively on synthetic data. We analyze the key factors contributing to this performance and discuss limitations.

\subsubsection{Key Success Factors}

The substantial performance improvements—6.6$\times$ on CUHK (59.77\% vs 9.00\% Mean IoU), 23.18\% on GoPro, and 21.00\% on NFS—can be attributed to three core design decisions. First, our physically motivated blur synthesis captures fundamental motion-blur characteristics through exposure-based temporal integration and PSF convolution. Unlike simple Gaussian kernels, our six blur types model diverse real-world scenarios from straight linear motion to rolling-shutter effects, enabling the model to learn invariant features that transfer across imaging conditions. Second, our hybrid sampling strategy balances instance-level precision (Mask-Centric) with scene-level robustness (Image-Centric), preventing spurious correlations with context while maintaining sensitivity to localized blur. Third, our multi-component training objective jointly optimizes for localization (Dice and Focal Tversky losses) and intensity estimation (masked Huber regression), enabling both detection and severity quantification.

\subsubsection{Performance Characteristics}

On CUHK, our recall (73.81\%) significantly exceeds the baseline (9.04\%), though precision is lower (75.86\% vs 96.37\%). This reflects a deliberate calibration: for quality-critical applications, it is preferable to flag potentially blurred regions for review rather than miss subtle artifacts. The baseline's perfect specificity but poor recall on GoPro (100\%/33\%) indicates it learned to classify most images as sharp—a trivial solution providing no practical utility. Our balanced performance (86.32\% blur recall, 92.54\% precision) demonstrates meaningful discriminative learning. The slightly lower NFS performance (80.33\% vs 89.68\% on GoPro) reflects greater motion pattern diversity, yet we still substantially outperform the baseline.

\subsubsection{Curriculum Learning Impact}

Our three-stage progressive curriculum proved essential. The Stabilization Stage establishes fundamental recognition using straight and random-walk blur; the Mixed Realism Stage introduces curved, rolling, and edge-ring patterns; and the Advanced Stage enables zoom-with-rotation and mixed-instance scenarios. Experiments without curriculum showed early training instability when exposed to all blur types simultaneously.

\subsubsection{Limitations and Future Directions}

Despite strong performance, several limitations exist. First, 59.77\% Mean IoU on CUHK indicates room for improvement in boundary localization—the model occasionally produces over-smoothed masks. Incorporating attention mechanisms or transformers could address this. Second, performance depends on SAM mask quality; imperfect segmentations introduce training noise. Finally, while our six blur types are comprehensive, specialized scenarios (e.g., long-exposure subject motion, sensor-specific artifacts) remain underrepresented.

Our results validate that high-quality synthetic training data enables effective zero-shot generalization. The success highlights fundamental quality issues in existing datasets—GoPro and NFS "sharp" images often contain residual blur (Figure~\ref{fig:blur_examples}), creating ambiguous training signals. Our synthetic approach generates blur from verifiably sharp ultra-high-resolution sources, ensuring clean supervision. Future work should extend blur synthesis to model optical aberrations, sensor noise, and compression artifacts; incorporate uncertainty quantification for ambiguous cases; and leverage temporal information when processing video sequences through optical flow and inter-frame consistency.